\documentclass[runningheads]{llncs}

\usepackage[utf8]{inputenc}
\usepackage[T1]{fontenc}
\usepackage{amsfonts}
\usepackage{amsmath}
\usepackage{hyperref}
\usepackage[capitalise]{cleveref}
\usepackage{placeins}
\usepackage{cite} 
\newcommand{\citep}{\cite} 
\usepackage{graphicx}
\usepackage{booktabs} 
\usepackage{adjustbox}
\usepackage[caption=false]{subfig}
\usepackage{microtype}
\usepackage{contour}

\usepackage{listings} 
\usepackage{xcolor}
\usepackage{pythonhighlight}

\lstnewenvironment{pythonLines}[1][]{\lstset{style=mypython,breaklines=false,basicstyle=\ttfamily\scriptsize,framexrightmargin=0em}}{}

\begin{document}

\title{QuickQual: Lightweight, convenient retinal image quality scoring with off-the-shelf pretrained models}

\titlerunning{QuickQual: Lightweight, convenient retinal image quality scoring}
\authorrunning{Engelmann et al.}

\makeatletter
\newcommand{\printfnsymbol}[1]{%
  \textsuperscript{\@fnsymbol{#1}}%
}
\makeatother

\author{Justin Engelmann\inst{1,2} 
\and Amos Storkey\inst{2}\thanks{Equal supervision.}
\and \\Miguel O. Bernabeu\inst{1,3}\printfnsymbol{1}}

\institute{Centre for Medical Informatics, Usher Institute, University of Edinburgh, Edinburgh,
Scotland, UK\\
\email{justin.engelmann@ed.ac.uk}
\and  Institute for Adaptive and Neural Computation, School of Informatics, University of Edinburgh, Edinburgh,
Scotland, UK
\and The Bayes Centre, University of Edinburgh, Edinburgh,
Scotland, UK}

\maketitle

\setcounter{footnote}{0}
\begin{abstract}

Image quality remains a key problem for both traditional and deep learning (DL)-based approaches to retinal image analysis, but identifying poor quality images can be time consuming and subjective. Thus, automated methods for retinal image quality scoring (RIQS) are needed. The current state-of-the-art is MCFNet, composed of three Densenet121 backbones each operating in a different colour space. MCFNet, and the EyeQ dataset released by the same authors, was a huge step forward for RIQS.
We present QuickQual, a simple approach to RIQS, consisting of a single ``off-the-shelf'' ImageNet-pretrained Densenet121 backbone plus a Support Vector Machine (SVM). QuickQual performs very well, setting a new state-of-the-art for EyeQ (Accuracy: 88.50\% vs 88.00\% for MCFNet; AUC: 0.9687 vs 0.9588). This suggests that RIQS can be solved with generic ``perceptual'' features learned on natural images, as opposed to requiring DL models trained on large amounts of fundus images. Additionally, we propose a Fixed Prior linearisation scheme, that converts EyeQ from a 3-way classification to a continuous logistic regression task. For this task, we present a second model, QuickQual MEga Minified Estimator (QuickQual-MEME), that consists of only 10 parameters on top of an off-the-shelf Densenet121 and can distinguish between gradable and ungradable images with an accuracy of 89.18\% (AUC: 0.9537). 
\href{https://github.com/justinengelmann/QuickQual}{Code and model are available on GitHub.} QuickQual is so lightweight, that the entire inference code (and even the parameters for QuickQual-MEME) is already contained in this paper.
  
\end{abstract}
\section{Introduction}

Retinal colour fundus images are used in ophthalmology for detecting and grading of retinal diseases like diabetic retinopathy, and also capture a detailed picture of the blood vessels, which could be informative about systemic health \citep{macgillivray2014retinal,wagner2020insights,velasco2021decreased}.. However, image quality is a key problem even when data is specifically collected for research purposes. For example, in UK Biobank, recent studies discarded 26\% \citep{zekavat2022deep} to 43\% \citep{velasco2021decreased} of the available images due to quality issues and only about 60\% of participants were found to have at least one good quality image \citep{macgillivray2015suitability}. However, Retinal Image Quality Scoring (RIQS) can be subjective and even graders with medical backgrounds only have moderate to substantial agreement \citep{laurik2022assessment}. Thus, automated RIQS methods are needed to provide objective and reproducible quality scores. Reproducibility is especially as image quality-based exclusions can introduce selection bias by excluding older, male, less-healthy, and non-White subjects more frequently \citep{engelmann2023exclusion}. Even work that develops retinal image improvement \citep{shen2020modeling} or robust retinal image analysis methods \citep{engelmann2022robust} depends on reliable quality scores.

\cite{fu2019evaluation} introduced an automated RIQS method called MultiColourspaceFusionNetwork (MCFNet) and the EyeQ dataset, a re-annotation of the publicly available Kaggle Diabetic Retinopathy dataset that provides quality annotations on a 3 class scale (Good, Usable, Reject). This work was a huge step forward for the field of RIQS with both MCFNet and the EyeQ dataset being very important contributions in their own right. The authors made the code, model weights, and data annotations publicly available, enabling others to both use and build on their work. However, MCFNet requires specific colourspace data transformation steps and consists of 3 Densenet121 backbones. Thus, MCFNet requires a specific dataloader and  model weights, and is a somewhat large model. Recent work showed that ``off-the-shelf'' DL models pretrained on ImageNet might be able to capture salient information such as age from retinal fundus images even without fine-tuning \citep{engelmann2023deep}. Inspired by that, we set out to investigate whether we can develop a simpler yet effective automated RIQS method that uses such an off-the-shelf model with a classical machine learning classifier.

Our main contributions are:
\begin{itemize}
    \item \textbf{QuickQual}, a simple RIQS method based on an ``off-the-shelf'' Densenet121 and an SVM, that achieves state-of-the-art on EyeQ while requiring only standard libraries and 14 lines of code;
    \item \textbf{Fixed Prior linearisation}, a simple method for converting EyeQ into a continuous task while retaining information about the Usable class;
    \item \textbf{QuickQual-MEME}, an even simpler version of QuickQual with a linear layer instead of an SVM that produces a continuous quality score. In fact, QuickQual-MEME is so lightweight, that the entire code and model parameters are contained in \cref{fig:QuickQualMemeCode}.
\end{itemize}

\section{Methods}
\subsection{EyeQ dataset}
We use the EyeQ dataset introduced by \cite{fu2019evaluation}, which provides quality annotations for a subset of the EyePacs Diabetic Retinopathy dataset on Kaggle, with three classes: Good, Usable, Bad. We preprocess the images by removing black areas and then padding the images to square in case they would be non-square otherwise.

\begin{figure}[t]
     \centering
    \includegraphics[width=0.85\textwidth]{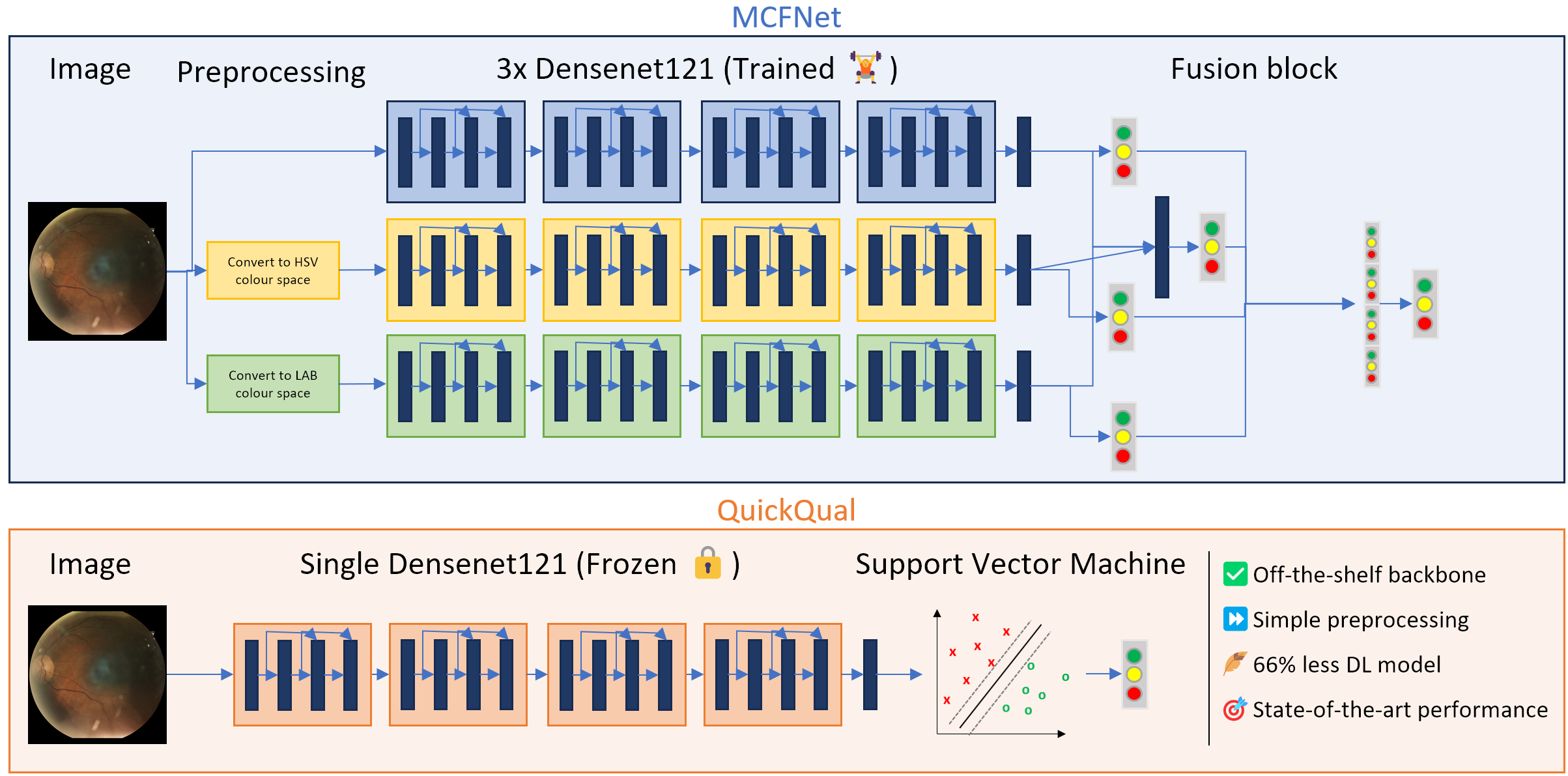}
    \caption{Comparison between MCFNet (top) and QuickQual (bottom). QuickQual-MEME uses a linear layer instead of SVM.}
    \label{fig:fig1overview}
\end{figure}
\subsection{QuickQual}
With QuickQual, we aim to develop a method that is quick and convenient to use. By that, we do not merely mean processing speed but also ease of implementation. Our goal is that with less than 20 lines of code and only standard Python libraries, a researcher could apply this method to their own images to obtain quality scores. Thus, we avoid complex preprocessing schemes and non-standard DL architecture code. We use a pretrained DL model from a standard Python DL library and instead of fine-tuning this on the EyeQ dataset, we simply keep it fixed and learn a Support Vector Machine (SVM) on top.

To enable an easier comparison with MCFNet, we also use Densenet121 \citep{huang2017densely} as our DL model, but with pre-trained ImageNet \citep{deng2009imagenet} weights from the pytorch image models (timm) \citep{rw2019timm} library. We use a SVM from scikit-learn with standard parameters, except setting ``probability=True'' to obtain probability scores from the SVM. To obtain discrete class labels, we take the class with the highest probability. We process images at a resolution of $512\times512$ and simply normalise all channels with mean and standard deviation parameters of $0.5$.

\subsection{RIQS beyond 3-way classification: Fixed prior linearisation}
In practice, individual probabilities for three separate classes can be inconvenient to use. Thus, previous work \citep{zhou2022automorph} focused on the binary task Gradable (Good or Usable) vs Bad (Reject) instead. This produces a single, continuous score where a simple cut-off for excluding images can be selected. However, this approach treats Good images exactly the same as Usable ones, losing the information that Usable images are at least slightly poorer quality. To remedy this, we propose a simple linearisation scheme with a fixed prior, i.e. that Usable images are in-between Good and Bad images in terms of quality. During model fitting, we set the optimal output p(Bad) that minimises the loss function to be $0$ for Good images, $1$ for Bad images, and our fixed prior $p$ for Usable images. In present work, we simply set $p=0.5$ and thus ask our model to map Usable images in-between Good and Bad ones, thus retaining the information in the labels. This should produce a smooth and desirable quality score but might reduce accuracy for the binary task.

\subsection{QuickQual MEga Minified Estimator (QuickQual-MEME)}
QuickQual-MEME is an even more lightweight, easy-to-use RIQS model consisting of a pretrained Densenet121 and only 10 parameters for a linear layer. QuickQual-MEME only needs standard python libraries and 15 lines of code. Instead of a SVM, QuickQual-MEME uses a Logistic Regression (Logit) with 10 parameters (9 weights, 1 bias) as classifier. To find these parameters, we proceeded as follows: First, we fit a Logit on the whole EyeQ training set with an L1 penalty (``Lasso'') with the default regularisation $C=1$ and the SAGA optimiser. We then examined the histogram of absolute coefficient magnitudes and chose a cut-off of $0.2$ to select 288 of the 1,024 Densenet121 variables. Next, we did forward step-wise features selection using 2-fold crossvalidation on the training set and the AUC as criterion to select the 9 most useful features. Finally, we rounded the parameters to two decimal places so they are easier to report and copy, which led to an insubstantial change in accuracy.

\subsection{Evaluation}
\label{sec:methods_eval}
For the standard EyeQ 3-way classification task, we use standard metrics like Accuracy, F1 score, area under the receiver operating characteristic curve (AUC), logistic loss also known as cross-entropy (LogLoss), cohen's unweighted Kappa (Kappa) and quadratic weighted Kappa (QuadKappa). AUC is a ranking metric that evaluates the model across all possible decision thresholds, whereas LogLoss provides a measure of calibration. Kappa captures how well the model agress with the labels compared to random chance, and QuadKappa penalises errors by more than one class much more, i.e. confusing Good with Bad is worse than confusing Good with Usable. For the binary Gradable vs. Ungradable, we use the same metrics except for Kappa/QuadKappa, which are only suitable for multi-class problems. We calculate all metrics using scikit-learn and use the predicted probabilities for MCFNet provided by the authors to ensure a fair and accurate comparison.\footnote{Note that for MCFNet, the original accuracy scores provided were not entirely accurate due to a bug in the evaluation code. See the note here on the Github for MCFNet: \url{https://github.com/HzFu/EyeQ\#-reference} ``\textit{Note: The corrected accuracy score of MCF-Net is 0.8800.}'' We thank the authors of MCFNet for their exceptional transparency in sharing not just code, model weights and data, but also their model's test set predictions.}

\section{Results}
\subsection{QuickQual performance on EyeQ}

\begin{table}[t]%
\caption{Performance for MCFNet and QuickQual on the test set of EyeQ (n=16,249). Note: All metrics are calculated from per-sample predictions using identical code to ensure an accurate comparison. See \cref{sec:methods_eval} and footnote 1.}
\label{tab:results}
\centering
\begin{adjustbox}{max width=0.95\textwidth, max totalheight=1\textheight-2\baselineskip}
{\small


\begin{tabular}{lccccccc}
\toprule

Model \hspace{4em} & Accuracy & AUC & F1 & LogLoss & Kappa & QuadKappa & Filesize\\
\midrule
MCFNet \citep{fu2019evaluation} & 0.8800 & 0.9588 & 0.8606 & 0.3632 & 0.8017 & 0.8955 & 112MB\\
QuickQual (ours) \hspace{0.3em} & \hspace{0.3em} \bfseries 0.8863 \hspace{0.3em} & \hspace{0.3em} \bfseries 0.9687 \hspace{0.3em} & \hspace{0.3em} \bfseries 0.8675 \hspace{0.3em} & \hspace{0.3em} \bfseries 0.3049 \hspace{0.3em} & \hspace{0.3em} \bfseries 0.8107 \hspace{0.3em} & \hspace{0.3em} \bfseries 0.9019 \hspace{0.3em} & \hspace{0.3em} \bfseries 31+25=56MB\\
\bottomrule
\end{tabular}
}
\end{adjustbox}
\end{table}%

\begin{figure}[t]
     \centering
    \includegraphics[width=0.75\textwidth]{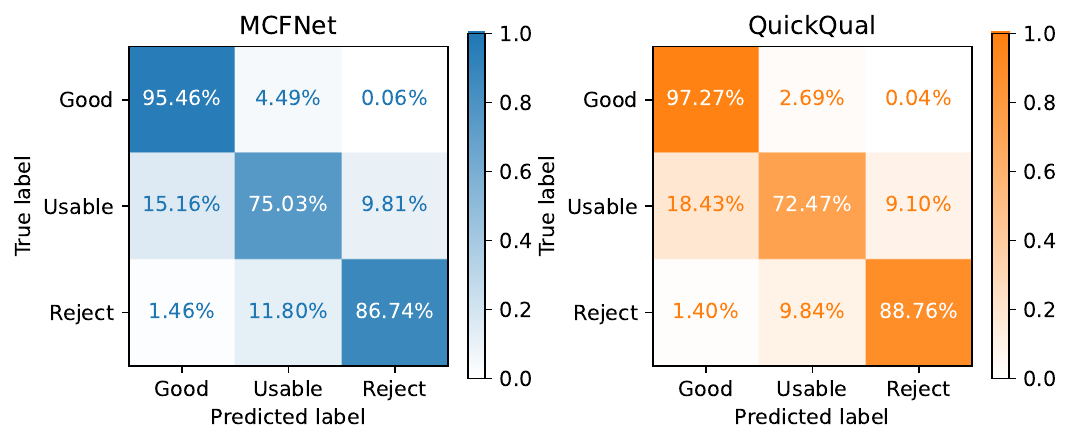}
    \caption{Confusion matrices for MCFNet and QuickQual, normalised per row.}
    \label{fig:confmatrix}
\end{figure}

\cref{tab:results} shows the results for QuickQual and MCFNet. QuickQual performs better in every metric. Accuracy, F1 and QuadKappa are slightly better, whereas AUC, LogLoss and Kappa are substantially better. QuadKappa penalises large errors (i.e. confusing Good with Reject) more than Kappa. Thus, QuickQual having a larger improvement in Kappa than in QuadKappa suggests that it is particularly good at distinguishing between the Usable and Good/Reject classes. The confusion matrix (\cref{fig:confmatrix}) shows that QuickQual is also better at avoiding large errors (top right and bottom left corners). The only category where QuickQual makes more errors than MCFNet is confusing Usable with Good (middle left). In our opinion, this error is the least concerning type of error - in fact previous work has even combined these two categories  \citep{zhou2022automorph}.

\begin{figure}[t]
     \centering
    \includegraphics[width=0.9\textwidth]{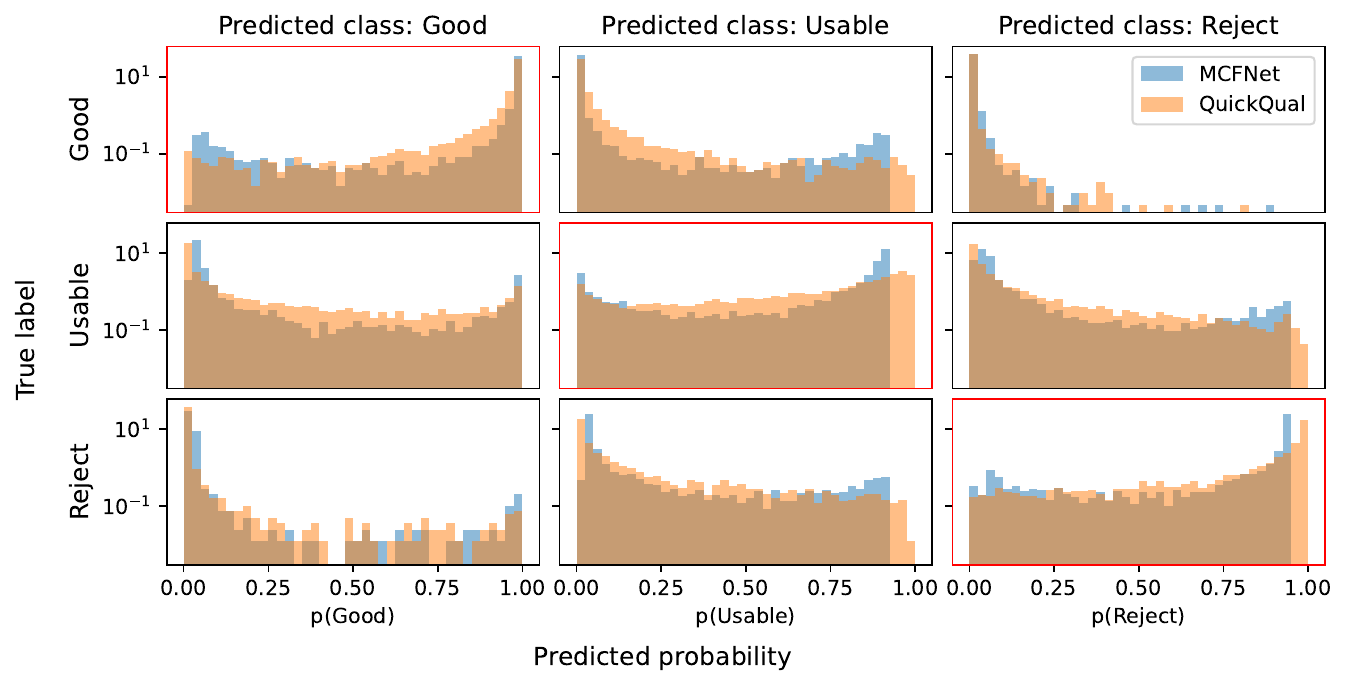}
    \caption{Distributions of predictions on EyeQ test set for each class, stratified by ground-truth class. Note that y-axis is on a log-scale. This plot is a ``soft'' version of a confusion matrix. For the diagonal (highlighted red) plots, predictions closer to 1 are better; whereas for the off-diagonal plots, predictions closer to 0 are better.}
    \label{fig:mcfnet_vs_quickqual_histograms}
\end{figure}

LogLoss is the metric with the largest difference, suggesting that QuickQual is much better calibrated. \cref{fig:mcfnet_vs_quickqual_histograms} shows the distributions of predicted probabilities for both models. Interestingly, MCFNet - unlike QuickQual - never predicts the Usable or Reject classes with large confidence. This might be a by-product of class imbalance and batch training. The QuickQual approach projects the images to small 1,024 dimensional vectors first, which then allows us to fit the SVM to all training images at once.

\subsection{QuickQual-MEME performance on binary task}

\begin{table}[t]%
\caption{Performance for binary task Gradable (Good/Usable) vs. Ungradable (Reject).}
\label{tab:results_bin}
\centering
\begin{adjustbox}{max width=0.7\textwidth, max totalheight=1\textheight-2\baselineskip}
{\small
\begin{tabular}{lcccc}
\toprule
 & Accuracy & AUC & F1 & LogLoss \\
\midrule
MCFNet \citep{fu2019evaluation} (Using p(Reject)) & 0.9459 & 0.9819 & 0.8640 & 0.1445 \\
QuickQual (Using p(Reject)) & \bfseries 0.9520 & \bfseries 0.9870 & \bfseries 0.8799 &  \bfseries 0.1162 \\
\midrule
QuickQual-MEME & 0.8918 & 0.9537 & 0.7602 & 0.2742 \\
QuickQual-Binary & 0.9404 & 0.9787 & 0.8505 & 0.1650 \\
\bottomrule
\end{tabular}

}
\end{adjustbox}
\end{table}%

\begin{figure}[t]
     \centering
    \includegraphics[width=0.7\textwidth]{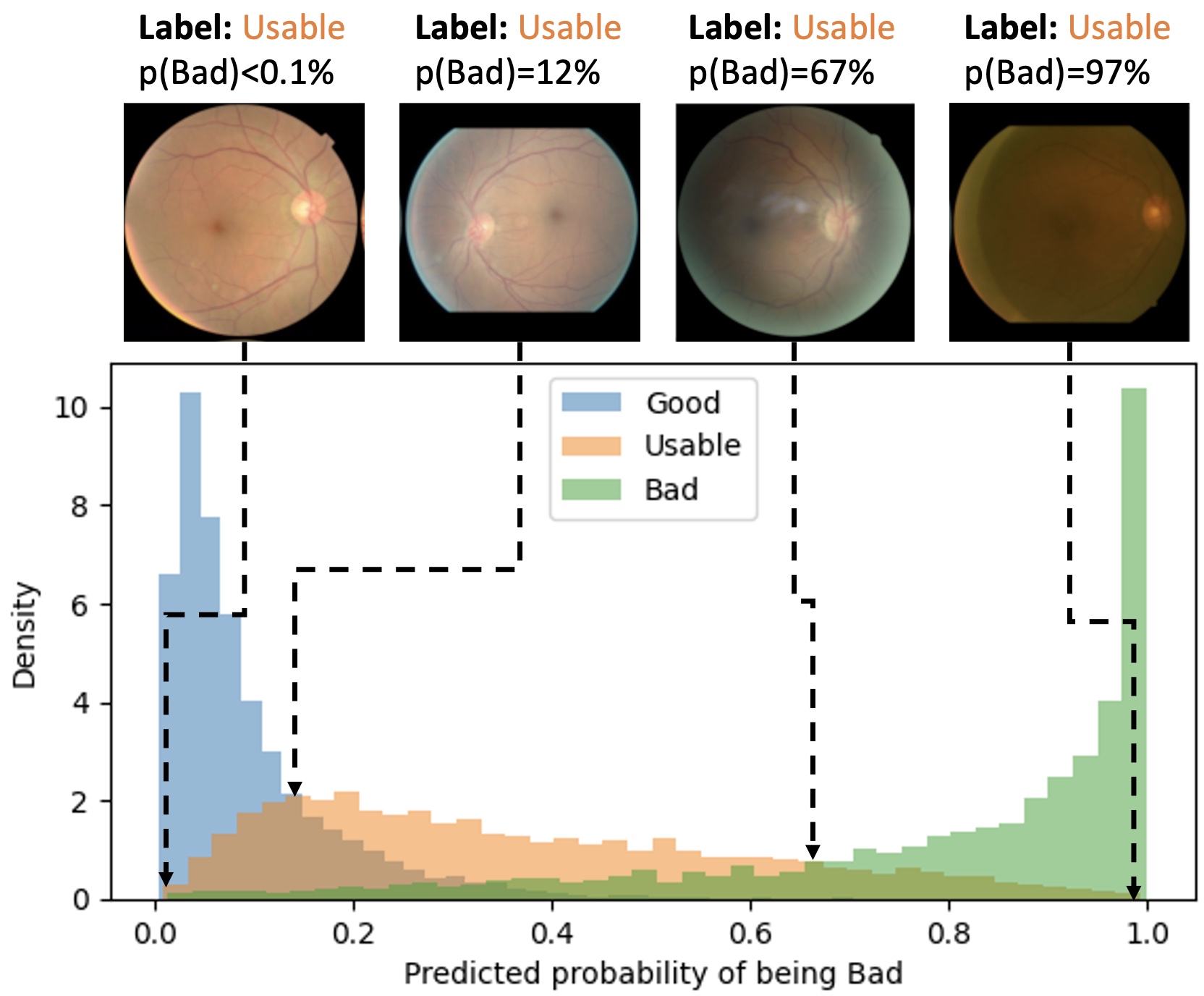}
    \caption{QuickQual-MEME predicted p(Bad) on the EyeQ test set, stratified by ground-truth class, with example images belonging to the Usable class shown above.}
    \label{fig:qqmeme_histogram}
\end{figure}

\cref{tab:results_bin} shows the results for the binarised task. For comparison, we also evaluate MCFNet and QuickQual on this task, using the predictions for the Reject class, as well as a QuickQual model trained on the binary task. The models trained on the original task perform best, with QuickQual offering slightly better performance in terms of Accuracy and AUC, and a large improvement for F1 and LogLoss over MCFNet. As expected, QuickQual-Binary using the SVM and all 1,024 Densenet121 features outperforms QuickQual-MEME which only uses 9 features. 

Interestingly, QuickQual-Binary is outperformed by QuickQual trained on the original task, suggesting that the fixed Prior Linearisation scheme reduces accuracy for Bad vs Good/Usable. However, QuickQual-MEME produces very smooth and desirable quality scores (\cref{fig:qqmeme_histogram}): The Good and Bad classes have modes on either extremes, while the Usable class is smoothly distributed in-between, with a mode closer to the Good class. This matches the class names: Usable is conceptually closer to Good than to Bad. Images from the Usable class with very low p(Bad) appear to be good quality, while those with high p(Bad) appear poor. Where the distributions of Good and Usable overlap, images are imperfect but still generally good; and where Usable and Bad overlap, they are poor. 

Although this evaluation is not comprehensive, this suggests QuickQual-MEME's quality score for the Usable class might align well with actual quality. Giving a very low p(Bad) score to all the Usable images, including the ones that look quite poor, would increase accuracy on the binarised task. However, in our opinion, the current behaviour of QuickQual-MEME appears preferable to that. Thus, accuracy might be an imperfect measure and more fine-grained expert evaluation is needed.

\begin{figure}[tb]
\begin{adjustbox}{max width=0.7\textwidth, max totalheight=1\textheight-2\baselineskip}
{
\begin{pythonLines}
import torch
from torchvision.transforms import functional as F
from PIL import Image
import timm
img = Image.open('[DATAFOLDER]/10036_left.jpeg')
model = timm.create_model('densenet121.tv_in1k', 
                          pretrained=True, num_classes=0)
model.eval().cuda()
w = torch.tensor([-1411.32, 517.09, 342.41, -707.9,
                  1442.09, -23.25, -541.64, -8.44, 5.44])
b = torch.tensor([5.18])
img = F.to_tensor(F.resize(img, 512))
img = F.normalize(img, [0.5]*3, [0.5]*3).cuda().unsqueeze(0)
with torch.no_grad():
    feats = model(img).squeeze().cpu().reshape(1, -1)
feats = feats[:, [71, 109, 121, 53, 55, 123, 29, 133, 84]]
pred = torch.sigmoid(feats @ w + b)
\end{pythonLines}
}
\end{adjustbox}
\caption{Entire inference code to run QuickQual-MEME, including the model parameters themselves. The code can be copied from the figure above.}
\label{fig:QuickQualMemeCode}
\end{figure}

\subsection{Convenience and speed}

\begin{figure}[tb]
     \centering
    \includegraphics[width=0.7\textwidth]{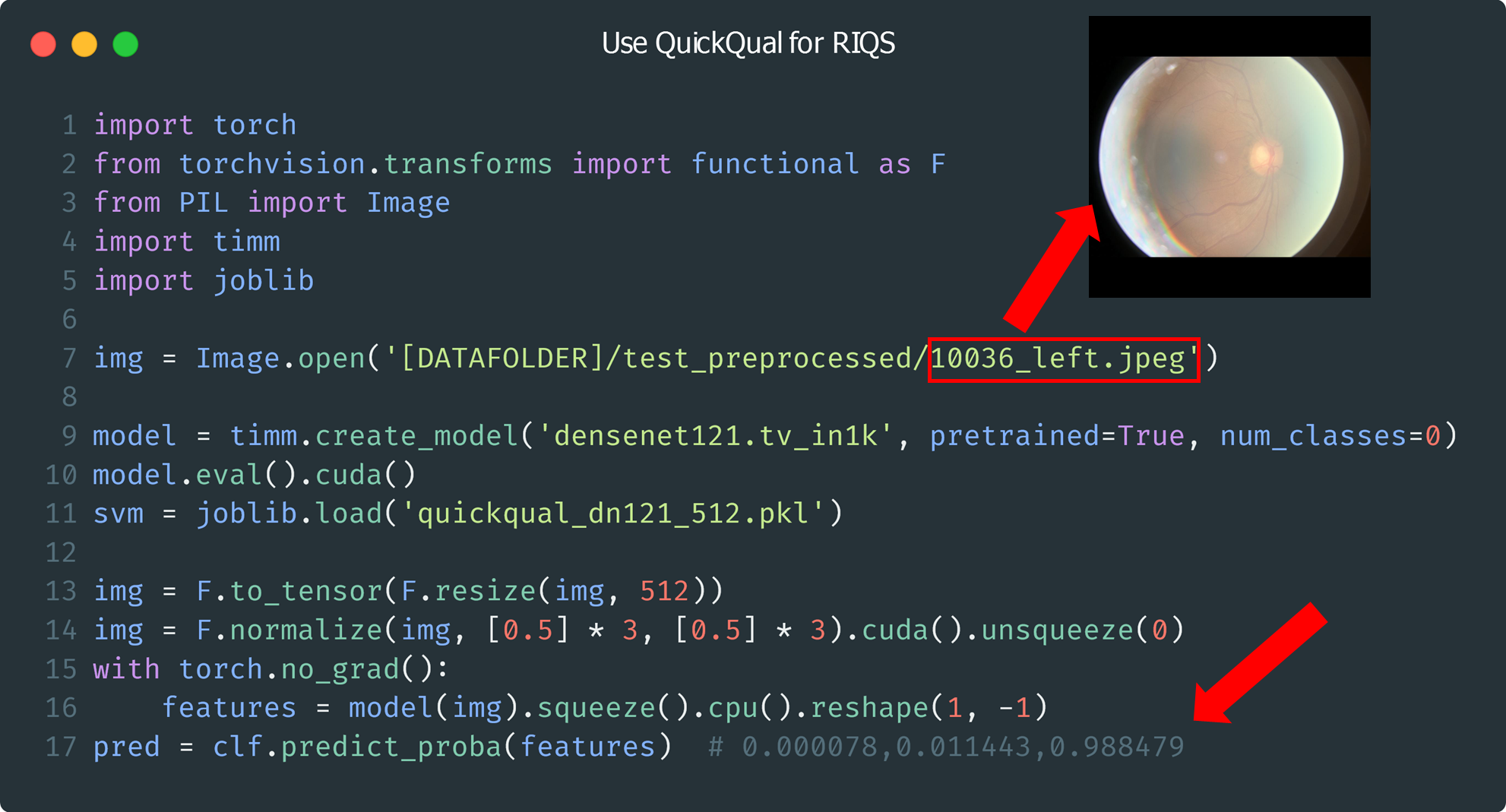}
    \caption{Entire inference code needed for QuickQual. Arrows highlight the example image which is of poor quality; and the prediction for p(Bad)$\approx 99\%$.}
    \label{fig:QuickQualInferenceExample}
\end{figure}

QuickQual and QuickQual-MEME need about 15 lines of code to be used together with standard, widely used libraries like PyTorch, scikit-learn and timm. QuickQual-MEME only need 10 parameters to be used, QuickQual needs a 25MB pretrained scikit-learn SVM. This means that QuickQual is very easy to implement and thus very convenient to use for researchers.

Inference times for a single images were measured across 1,000 repetitions, with times reported being mean and standard deviation. QuickQual processed the image in 16.6 ms ± 602 µs on a GPU and 79.5 ms ± 2.45 ms on a CPU. QuickQual-MEME took 14.5 ms ± 536 µs on a GPU and 79.3 ms ± 1.88 ms on a CPU. These times suggest that the SVM only adds minimal overhead compared to a linear model when the Densenet121 is GPU-accelerated and no noticable overhead when no GPU is used. Note that batched inference for multiple images in parallel will likely be even faster per image, but even when processing images one-by-one, 767 images could be processed per minute on a CPU. A time of less than a tenth of a second on a CPU also means that QuickQual could conceivably be deployed in practice to assess images in real time as they are taken.

\FloatBarrier
\section{Discussion}

We presented QuickQual, which achieves state-of-the-art on EyeQ with only 14 lines of inference code, and QuickQual-MEME which produces a single continuous quality score and fits in \cref{fig:QuickQualMemeCode}. We hope that these will be an easy-to-use, convenient method for other researchers in the field.

We also introduced a Fixed Prior linearisation scheme that better preserves information about the Usable class. While quantitatively this reduced accuracy, limited qualitative evaluation suggests that it might produce a smooth, desirable quality score. 

In the future, we plan to evaluate this in more detail by having experts rank images in terms of quality and examining the correlation with QuickQual-MEME's quality score. We also plan evaluate other pretrained DL models to see whether a similarly performant yet more light-weight model could be found that enables even faster computation of quality scores. Finally, we plan to externally validate QuickQual and QuickQual-MEME on images from UK Biobank.

\section*{Acknowledgements}
We thank our friends and colleagues for their help and support. J.E. and this work was supported by the United Kingdom Research and Innovation (grant EP/S02431X/1), UKRI Centre for Doctoral Training in Biomedical AI at the University of Edinburgh, School of Informatics.  For the purpose of open access, the author has applied a creative commons attribution (CC BY) licence to any author accepted manuscript version arising.

\FloatBarrier

\bibliographystyle{splncs04}
\bibliography{references.bib}

\end{document}